\documentclass{article}
\usepackage{spconf,amsmath,amssymb,graphicx,hyperref,booktabs}
\hypersetup{hidelinks,pdfauthor={Fred Sun, Jingze Wang, Minkun Xu, Shangqi Guo}}

\title{Time-Aligned Evolving Concept Graphs\\for Scientific Relation Forecasting}
\name{Fred Sun$^{1,*}$, Jingze Wang$^{1,*}$, Minkun Xu$^{2}$, Shangqi Guo$^{1,\dagger}$
\thanks{$^{*}$Equal contribution. $^{\dagger}$Corresponding author.}}
\address{$^{1}$Center for Brain-Inspired Computing Research,\\
Department of Precision Instrument, Tsinghua University, Beijing, China\\
$^{2}$Guangdong Institute of Intelligence Science and Technology, Zhuhai, China}

\begin{document}
\maketitle
\begin{abstract}
Forecasting scientific relations can guide discovery by identifying promising connections before they emerge. Existing approaches often model concept semantics and graph structure separately or summarize semantics over coarse historical snapshots, leaving semantic representations potentially misaligned with rapidly evolving graph evidence. We propose a time-aligned evolving concept graph framework that jointly models semantic and structural evolution. Its core idea is to treat dated papers as shared update events, reconstructing semantic and structural states from the same publication history through each prediction time. Pair-level fusion combines these states to forecast first co-occurrence, relation formation, and conditional relation type. Holding architecture and training fixed, refreshing context alongside graph updates improves mean relation AUPRC by 16.6\% over frozen context. On a graph built from 187,848 papers with 270,687 concepts and 7.45 million co-occurrence links, the complete framework improves mean relation AUROC from 0.9290 for the strongest evaluated baseline to 0.9722, with mean population-weighted AUPRC 0.005778.
\end{abstract}


\section{Introduction}\label{sec:intro}
Scientific advances often connect separate lines of work: a method enters a new application, techniques are combined, or evidence challenges an established approach. Forecasting these connections can reveal promising directions before findings appear. Concept-graph methods predict future co-occurrences from historical literature~\cite{krenn2020predicting,krenn2023forecasting,marwitz2026predicting}. A scientific relation gives a connection a specific meaning, such as one method \emph{using}, \emph{combining with}, or \emph{replacing} another. Co-discussion alone does not establish these roles. We therefore forecast both whether a scientific relation will form and its type. These forecasts draw on concept usage and observed connections. When a method enters a new application, its recent contexts and emerging graph neighborhood describe the same development. Combining earlier concept usage with newer graph events mixes different stages of that development. Both sources should therefore be reconstructed at the same prediction time.

\begin{figure*}[t]
\centering
\includegraphics[width=\textwidth]{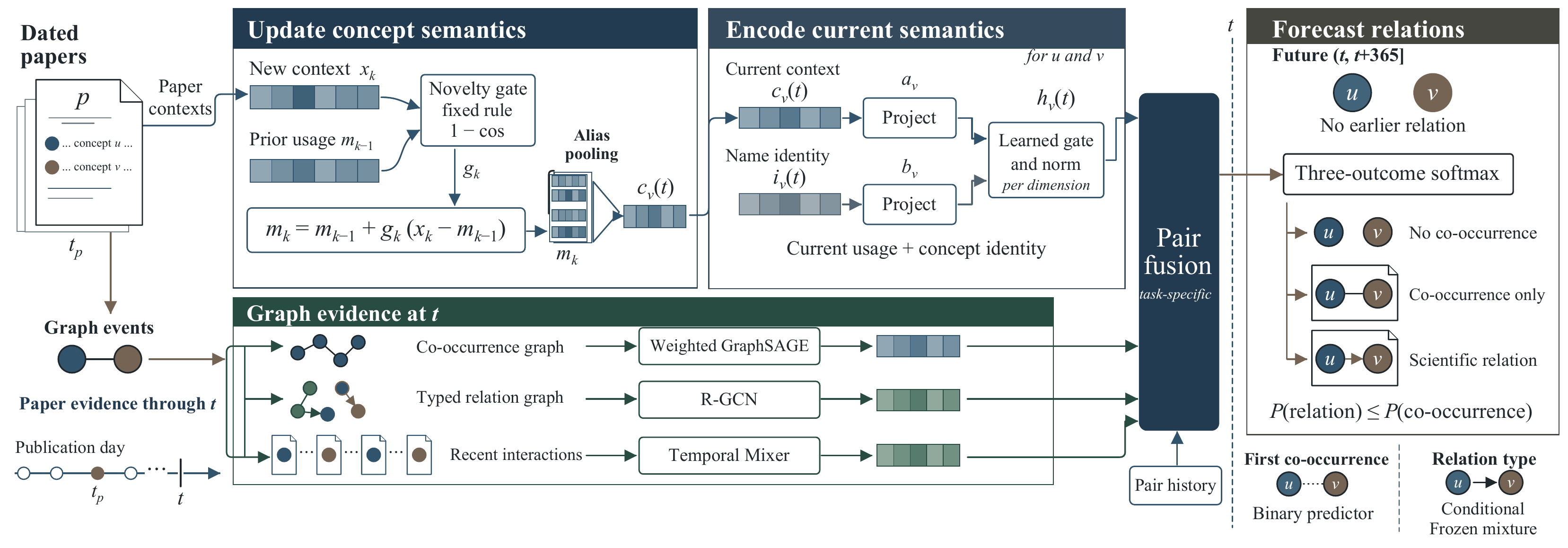}
\caption{Time-aligned evolving concept graphs. Dated papers provide contextual occurrences and graph events through query time $t$. Reconstructed semantic and structural states are fused at the concept-pair level to forecast first co-occurrence, relation formation, and conditional relation type.}
\label{fig:overview}
\end{figure*}

Literature-based discovery and scientific embeddings provide evidence for unobserved connections~\cite{swanson1986undiscovered,tshitoyan2019unsupervised}. Marwitz et al.~\cite{marwitz2026predicting} average contextual embeddings across abstracts before a cutoff year and combine them with graph information for link prediction. Temporal graph models~\cite{rossi2020temporal,xu2020inductive,cong2023graphmixer,yu2023dygformer} and text-attributed dynamic graphs~\cite{zhang2024dtgb} motivate the representation of evolving interactions. Diachronic embeddings~\cite{hamilton2016diachronic} examine semantic change, while temporal knowledge-graph models forecast future relations~\cite{jin2020renet}.

We propose a time-aligned evolving concept graph framework (Fig.~\ref{fig:overview}). Dated papers serve as shared update events for concept usage and graph connections. At each query, semantic and structural states are reconstructed through the same publication date, pairing current concept usage with its corresponding graph evidence. A context-sensitive update tracks usage; graph encoders and a recent-event pathway capture established connections and current activity. Pair-level fusion forecasts first co-occurrence, relation formation, and conditional relation type. With architecture and training held fixed, refreshing semantic context alongside graph updates improves mean relation AUPRC by $16.6\%$ over frozen context. The complete framework improves relation population-AUPRC by $70.7\%$ over the strongest evaluated baseline.

\section{Proposed Framework}\label{sec:method}
\subsection{Dated papers as shared update events}\label{sec:evidence}
\looseness=-1
We extract concept mentions from abstracts and consolidate aliases through retrieval and verification. The concept inventory extends chronologically: newly observed names receive existing or new node IDs, while prior assignments remain fixed. A corpus-wide acronym-ambiguity check filters candidate merges. Each paper supplies contextual occurrences, co-occurrences, and full-text evidence for typed, directed relations: \emph{uses}, \emph{combines}, \emph{replaces}, and \emph{contradicts}. A generator--discriminator teacher pipeline distills local relation extractors. Student confidence is calibrated on the extractor development split. Trusted instances have teacher evidence, agreement of at least two student extractors, or confidence $\ge0.95$. Relation records retain the paper, date, type, direction, confidence, and supporting text. An expert audit of 10,000 accepted relation instances yielded 93.4\% relation validity, 90.7\% type correctness, and 89.9\% direction correctness.

\looseness=-1
For each query date $t$, reconstruction of semantic states and graph statistics uses no publications after that date. Only names observed by $t$ are activated, and papers published on $t$ are included. The co-occurrence graph aggregates supporting-paper counts; the relation graph preserves types, directions, and evidence strength. We filter paper events by the query date before accumulating edge weights and relation statistics.

We forecast a first trusted scientific relation during the following-year window $(t,t+H]$; $H=365$ days for the test window. Eligible pairs have no prior scientific relation and involve only concepts observed by $t$; prior co-occurrence is allowed. A negative label indicates no observed relation during the window. Queries without complete future observation are excluded. \emph{First co-occurrence forecasting} considers pairs never previously observed together. \emph{Relation type prediction} classifies the first relation of a relation-positive pair. We merge \emph{replaces} and \emph{contradicts} and exclude conflicting first-day categories. The graph retains direction; formation and type targets are invariant to endpoint order. Within the window, relations entail co-occurrence; type is conditional on formation.

\subsection{Reconstructing concept semantics}\label{sec:semantic}
For each concept surface form in a paper, SciDeBERTa~\cite{jeong2022scideberta} provides contextual token representations. Averaging over the concept's tokens and repeated mentions yields one 768-dimensional occurrence vector. The retained record contains the surface form, paper identifier, publication day, and vector. 

To combine accumulated usage with recent adaptation, we initialize the state from the mean of earlier occurrences and sequentially update over the latest $K=20$ visible occurrences. For histories of at most $K$ occurrences, the first initializes the state and the remainder are replayed. Let $x_k$ denote the $k$-th visible occurrence, with $k$ counting the full history. Each replay step is
\begin{align}
m_k &= m_{k-1}+g_k(x_k-m_{k-1}),\label{eq:update}\\
g_k &= \max\{1/k,\ \sigma(\alpha[1-\cos(m_{k-1},x_k)]+\beta)\}.
\nonumber
\end{align}
The $1/k$ averaging rate provides a lower bound, emphasizing early observations while decreasing as history grows. The sigmoid term maintains a persistent response to new contexts, with $\beta$ setting the baseline and $\alpha$ controlling sensitivity to contextual discrepancy. We fix $\alpha=4$ and $\beta=-2.5$: the sigmoid rate is $0.076$ at zero cosine discrepancy and $0.109$ at similarity $0.9$. Multiplication by $x_k-m_{k-1}$ makes updates small when observations already resemble the state, while departures from accumulated usage receive more weight.

The historical mean retains older evidence, while recent replay adapts to current usage with at most $K$ sequential updates per surface form. At the test cutoff, $96.7\%$ of surface forms with contextual observations have at most 20 occurrences, so their histories are replayed in full. Reconstructing the state at each query ties semantic adaptation to the graph's publication cutoff.

We pool aliases observed by $t$ using their mention counts through $t$, then normalize to obtain contextual semantics $c_v(t)$. A separate 256-dimensional identity representation $i_v(t)$ averages embeddings of names observed by $t$. New contexts update usage, while new aliases can update identity.

\subsection{Forecasting from time-aligned states}\label{sec:prediction}
\emph{Contextual usage and concept identity.} Let $a_v=f_c(c_v(t))$ and $b_v=f_i(i_v(t))$ be learned projections. A gate combines their dimensions into the semantic state:
\begin{align}
\gamma_v &= \sigma(f_g([a_v,b_v])),\nonumber\\
h_v(t) &= \operatorname{Norm}\bigl(\gamma_v\odot a_v+(1-\gamma_v)\odot b_v\bigr).
\label{eq:semantic}
\end{align}
The gate balances concept identity and current usage in each feature dimension. This semantic state enters pair-level fusion alongside graph evidence from the same query date.

\emph{Accumulated structure and recent interactions.} Two layers of weighted GraphSAGE~\cite{hamilton2017inductive} encode co-occurrence neighborhoods. Two layers of basis-decomposed relational graph convolution~\cite{schlichtkrull2018modeling} encode typed relation neighborhoods, retaining direction and relation-specific evidence. 

Each endpoint also contributes its latest 20 co-occurrence and relation events. Tokens contain event type, age, intensity, and validity, with fixed time encodings. Two temporal MLP-Mixer layers~\cite{tolstikhin2021mlpmixer,cong2023graphmixer} summarize the sequence. This pathway describes recent activity alongside the accumulated neighborhoods.

\emph{Fusion and hierarchical outputs.} For endpoint embeddings $a,b$, we use $[a+b,|a-b|,a\odot b]$ to form an order-invariant pair representation. Task-specific gates combine semantic, graph, and event representations with pair-history statistics. The relation score contains a base prediction and gated ranking and interaction residuals. These contributions enter the relation logit before the three-outcome softmax:
\begin{align}
(p_{\mathrm{none}},p_{\mathrm{co\ only}},p_{\mathrm{rel}})
 &=\operatorname{softmax}(0,z_{\mathrm{co}},z_{\mathrm{rel}}),\\
P(\text{co-occurrence})&=p_{\mathrm{co\ only}}+p_{\mathrm{rel}},\nonumber\\
P(\text{relation})&=p_{\mathrm{rel}}\le P(\text{co-occurrence}).
\label{eq:hierarchy}
\end{align}
The outcomes describe no co-occurrence, co-occurrence without a relation, and relation formation within the target window. Here co-occurrence includes renewed co-occurrence of a previously observed pair; the inequality holds by construction.

First co-occurrence uses a direct binary output and a gated event correction. Relation type uses a conditional classifier mixing two frozen semantic-and-relation predictors selected for ROC and precision--recall, with a bounded semantic and graph residual. These task-specific outputs respect their different candidate sets.

\subsection{Training for rare relation formation}\label{sec:training}
For relation formation, training combines population samples, positives paired with broadly sampled negatives, and high-scoring difficult negatives. Inverse-probability weights correct the sampled label distribution. The objective combines weighted three-outcome negative log-likelihood, co-occurrence ranking, relation ranking, and conditional relation binary cross-entropy, with respective weights $1$, $0.2$, $0.8$, and $0.1$. The ranking loss uses $\operatorname{softplus}(s^- - s^+)$ for positive--negative score pairs; gathering candidates across devices enlarges the ranking pool.

We train from random initialization and select checkpoints on validation. Within the Pareto frontier of AUROC and AUPRC, we retain the highest AUPRC subject to relation AUROC $\ge0.9700$. The selected checkpoint is frozen before testing.

\section{Experiments}\label{sec:experiments}

\begin{figure*}[t]
\centering
\includegraphics[width=0.9\textwidth]{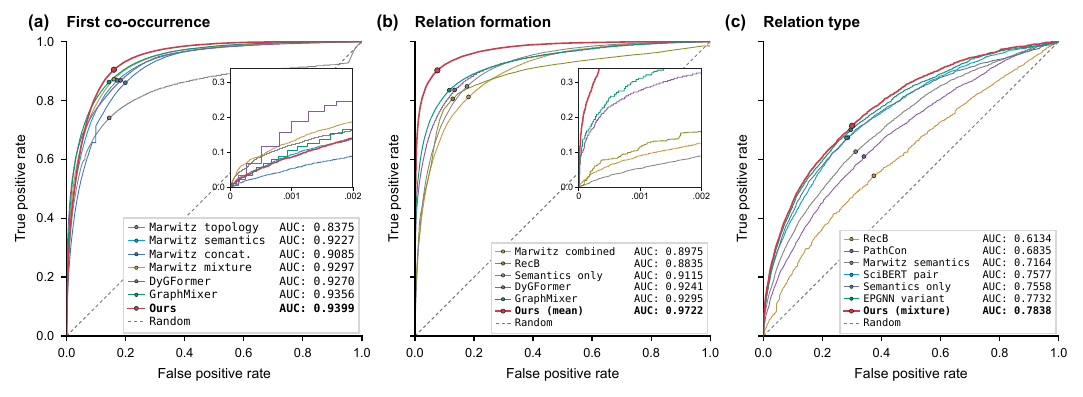}
\caption{ROC curves: (a,b) population weighted; (c) mean of three one-vs-rest curves. Our relation curve is the mean ROC from the three refreshed checkpoints in Tables~\ref{tab:main} and~\ref{tab:refresh}; other curves show representative runs. Insets magnify low false-positive rates.}
\label{fig:roc_overview}
\end{figure*}

\begin{table}[t]
\centering\small
\caption{Relation formation on the population-weighted test (mean $\pm$ sample SD, three seeds).}
\label{tab:main}
\vspace{6pt}
\setlength{\tabcolsep}{3pt}
\begin{tabular}{@{}lcc@{}}
\toprule
Model & AUROC & AUPRC ($\times10^{-3}$)\\
\midrule
Marwitz combined~\cite{marwitz2026predicting} & $0.8982\pm0.0064$ & $2.156\pm1.212$\\
DyGFormer~\cite{yu2023dygformer} & $0.9238\pm0.0016$ & $2.533\pm0.539$\\
GraphMixer~\cite{cong2023graphmixer} & $0.9290\pm0.0029$ & $3.384\pm1.600$\\
Ours (refreshed) & $\mathbf{0.9722}\pm0.0004$ & $\mathbf{5.778}\pm1.658$\\
\bottomrule
\end{tabular}
\end{table}

\subsection{Data, protocol, and comparisons}\label{sec:setup}
The computer-vision corpus contains 187,848 papers from January 2017 to June 2026, 270,687 concepts, 7,452,716 co-occurrence pairs, and 580,615 distinct pairs with scientific relations. We train on the 2022--2023 and 2023--2024 windows, validate on 2024--2025, and test on 2025--2026, with June 25 cutoffs. The test query is June 25, 2025; all methods share the concept inventory and use names and paper evidence observed through that date.

At the test cutoff, the population comprises 24.90 billion eligible concept pairs, with relation prevalence $1.9\times10^{-6}$. We form the test view from all target-window co-occurring pairs and uniformly sampled eligible pairs. From this view, we retain all 47,433 relation positives and uniformly sample 202,567 negatives. All models use this fixed 250,000-pair sample and inverse-probability weights accounting for both sampling stages. We report weighted AUROC and average precision (AUPRC). First co-occurrence uses its corresponding weighted view; relation type uses all 46,466 unambiguous relation-positive pairs and macro-averaged metrics.

Marwitz's combined predictor~\cite{marwitz2026predicting} provides a scientific-link comparison; DyGFormer and GraphMixer provide temporal graph comparisons. Figure~\ref{fig:roc_overview} also includes semantic and relational predictors, such as RecB~\cite{gastinger2024recb} and SciBERT pair embeddings~\cite{beltagy2019scibert}.

Our relation model uses hidden dimension 96, semantic and task widths 192, and dropout 0.2. AdamW runs for 1,600 steps with global batch size 24,576, learning rate $2\times10^{-4}$, 120 warmup steps, cosine decay, and weight decay 0.01. Validation runs every 200 steps. Both tables report means and sample standard deviations over seeds 17/23/42. The main result and refreshed row reuse identical checkpoints; the frozen comparison pairs the same seeds and training schedule.

\subsection{Forecasting scientific connections}\label{sec:performance}
The complete framework achieves relation AUPRC $0.005778$ and AUROC $0.9722$ (Table~\ref{tab:main}), improving AUPRC by $70.7\%$ over GraphMixer, the strongest evaluated baseline. The gain in population-weighted AUPRC reflects improved retrieval of rare future relations among previously unrelated concepts.

Figure~\ref{fig:roc_overview} compares ROC profiles across the three tasks. First co-occurrence achieves AUROC $0.9399$. For relation type, the frozen teacher mixture selected on validation reaches macro-AUROC $0.7838$ and macro-AUPRC $0.546491$. These complementary tasks distinguish detecting new shared usage from identifying the scientific role of a connection once it forms.

\begin{table}[t]
\centering\small
\caption{Context-refresh ablation for relation formation (mean $\pm$ sample SD, three paired seeds). Refreshed uses the Ours checkpoints from Table~\ref{tab:main}.}
\label{tab:refresh}
\vspace{6pt}
\setlength{\tabcolsep}{4pt}
\begin{tabular}{@{}lcc@{}}
\toprule
Context & AUROC & AUPRC ($\times10^{-3}$)\\
\midrule
Frozen & $0.9709\pm0.0007$ & $4.955\pm1.174$\\
Refreshed & $\mathbf{0.9722}\pm0.0004$ & $\mathbf{5.778}\pm1.658$\\
\bottomrule
\end{tabular}
\end{table}

\subsection{The value of refreshing concept semantics}\label{sec:refresh}
The refresh experiment tests whether updating concept semantics adds predictive value when graph evidence already follows the query date. The frozen variant retains each concept's context from the first benchmark query at which the concept is available. The refreshed variant reconstructs it at each query from the newly available paper history. Identity, graph, and event inputs follow the query date in both variants.

Both variants use the same architecture, fixed semantic update rule, replay budget, trainable modules, and training schedule. Only access to subsequent contexts changes, isolating the value of keeping semantic evidence current as the graph evolves. All six checkpoints are selected on validation and locked before the common population-weighted test.

Refreshing context raises relation population-AUPRC from $0.004955$ to $0.005778$, a $16.6\%$ improvement, and AUROC from $0.9709$ to $0.9722$ (Table~\ref{tab:refresh}). Both metrics improve in all three paired seeds. AUROC measures overall ranking; population-weighted AUPRC emphasizes precision when retrieving rare future relations. The frozen variant already receives new connections and recent interactions; refreshing context adds useful evidence about concept usage beyond these structural updates. The gain supports aligning concept usage with graph evidence at the prediction time.

\section{Conclusion}\label{sec:conclusion}
We presented a time-aligned evolving concept graph framework that uses dated papers as shared update events for semantic and structural states. With architecture and training held fixed, refreshing context yields $16.6\%$ higher AUPRC than frozen context, with both relation metrics improving across three paired seeds. Newly observed concept usage adds forecasting evidence beyond updated graph structure and recent interactions. The complete framework improves relation population-AUPRC by $70.7\%$ over the strongest evaluated baseline.

\pagebreak
\bibliographystyle{IEEEbib}
\bibliography{refs}

@article{marwitz2026predicting,
  title={Predicting new research directions in materials science using large language models and concept graphs},
  author={Marwitz, Thomas and Colsmann, Alexander and Breitung, Ben and others},
  journal={Nature Machine Intelligence},
  volume={8},
  pages={535--544},
  year={2026},
  doi={10.1038/s42256-026-01206-y}
}

@article{krenn2020predicting,
  title={Predicting research trends with semantic and neural networks with an application in quantum physics},
  author={Krenn, Mario and Zeilinger, Anton},
  journal={Proceedings of the National Academy of Sciences},
  volume={117},
  number={4},
  pages={1910--1916},
  year={2020},
  publisher={National Academy of Sciences}
}

@article{krenn2023forecasting,
  title={Forecasting the future of artificial intelligence with machine learning-based link prediction in an exponentially growing knowledge network},
  author={Krenn, Mario and Buffoni, Lorenzo and Coutinho, Bruno and Eppel, Sagi and Foster, Jacob Gates and Gritsevskiy, Andrew and Lee, Harlin and Lu, Yichao and Moutinho, Jo{\~a}o P. and Sanjabi, Nima and Sonthalia, Rishi and Tran, Ngoc Mai and Valente, Francisco and Xie, Yangxinyu and Yu, Rose and Kopp, Michael},
  journal={Nature Machine Intelligence},
  volume={5},
  number={11},
  pages={1326--1335},
  year={2023},
  publisher={Nature Publishing Group}
}

@inproceedings{beltagy2019scibert,
  title={{SciBERT}: A Pretrained Language Model for Scientific Text},
  author={Beltagy, Iz and Lo, Kyle and Cohan, Arman},
  booktitle={Proceedings of the 2019 Conference on Empirical Methods in Natural Language Processing (EMNLP-IJCNLP)},
  pages={3615--3620},
  year={2019}
}

@article{hamilton2017inductive,
  title={Inductive Representation Learning on Large Graphs},
  author={Hamilton, William L. and Ying, Rex and Leskovec, Jure},
  journal={Advances in Neural Information Processing Systems (NeurIPS)},
  volume={30},
  pages={1024--1034},
  year={2017}
}

@misc{rossi2020temporal,
  title={Temporal Graph Networks for Deep Learning on Dynamic Graphs},
  author={Rossi, Emanuele and Chamberlain, Ben and Frasca, Fabrizio and Eynard, Davide and Monti, Federico and Bronstein, Michael},
  year={2020},
  eprint={2006.10637},
  archivePrefix={arXiv}
}

@misc{xu2020inductive,
  title={Inductive Representation Learning on Temporal Graphs},
  author={Xu, Da and Ruan, Chuanwei and Korpeoglu, Evren and Kumar, Sushant and Achan, Kannan},
  year={2020},
  eprint={2002.07962},
  archivePrefix={arXiv}
}

@inproceedings{schlichtkrull2018modeling,
  title={Modeling Relational Data with Graph Convolutional Networks},
  author={Schlichtkrull, Michael and Kipf, Thomas N. and Bloem, Peter and van den Berg, Rianne and Titov, Ivan and Welling, Max},
  booktitle={The Semantic Web -- ESWC 2018},
  pages={593--607},
  year={2018}
}

@article{swanson1986undiscovered,
  title={Undiscovered public knowledge},
  author={Swanson, Don R.},
  journal={The Library Quarterly},
  volume={56},
  number={2},
  pages={103--118},
  year={1986},
  publisher={University of Chicago Press}
}

@article{tshitoyan2019unsupervised,
  title={Unsupervised word embeddings capture latent knowledge from materials science literature},
  author={Tshitoyan, Vahe and Dagdelen, John and Weston, Leigh and Dunn, Alexander and Rong, Ziqin and Kononova, Olga and Persson, Kristin A. and Ceder, Gerbrand and Jain, Anubhav},
  journal={Nature},
  volume={571},
  number={7763},
  pages={95--98},
  year={2019},
  publisher={Nature Publishing Group}
}

@inproceedings{jin2020renet,
  title={Recurrent Event Network: Autoregressive Structure Inference over Temporal Knowledge Graphs},
  author={Jin, Woojeong and Qu, Meng and Jin, Xisen and Ren, Xiang},
  booktitle={Proceedings of the 2020 Conference on Empirical Methods in Natural Language Processing (EMNLP)},
  pages={6669--6683},
  year={2020}
}

@inproceedings{cong2023graphmixer,
  title={Do We Really Need Complicated Model Architectures for Temporal Networks?},
  author={Cong, Weilin and Zhang, Si and Kang, Jian and Yuan, Baichuan and Wu, Hao and Zhou, Xin and Tong, Hanghang and Mahdavi, Mehrdad},
  booktitle={International Conference on Learning Representations (ICLR)},
  year={2023}
}

@inproceedings{yu2023dygformer,
  title={Towards Better Dynamic Graph Learning: New Architecture and Unified Library},
  author={Yu, Le and Sun, Leilei and Du, Bowen and Lv, Weifeng},
  booktitle={Advances in Neural Information Processing Systems (NeurIPS)},
  volume={36},
  year={2023}
}

@inproceedings{gastinger2024recb,
  title={History Repeats Itself: A Baseline for Temporal Knowledge Graph Forecasting},
  author={Gastinger, Julia and Meilicke, Christian and Errica, Federico and Sztyler, Timo and Schuelke, Anett and Stuckenschmidt, Heiner},
  booktitle={Proceedings of the Thirty-Third International Joint Conference on Artificial Intelligence (IJCAI)},
  pages={4016--4024},
  year={2024},
  doi={10.24963/ijcai.2024/444}
}

@inproceedings{hamilton2016diachronic,
  title={Diachronic Word Embeddings Reveal Statistical Laws of Semantic Change},
  author={Hamilton, William L. and Leskovec, Jure and Jurafsky, Dan},
  booktitle={Proceedings of the 54th Annual Meeting of the Association for Computational Linguistics (Volume 1: Long Papers)},
  pages={1489--1501},
  year={2016},
  address={Berlin, Germany},
  publisher={Association for Computational Linguistics},
  doi={10.18653/v1/P16-1141}
}

@article{jeong2022scideberta,
  title={{SciDeBERTa}: Learning {DeBERTa} for Science Technology Documents and Fine-Tuning Information Extraction Tasks},
  author={Jeong, Yuna and Kim, Eunhui},
  journal={IEEE Access},
  volume={10},
  pages={60805--60813},
  year={2022},
  publisher={IEEE},
  doi={10.1109/ACCESS.2022.3180830}
}

@inproceedings{zhang2024dtgb,
  title={{DTGB}: A Comprehensive Benchmark for Dynamic Text-Attributed Graphs},
  author={Zhang, Jiasheng and Chen, Jialin and Yang, Menglin and Feng, Aosong and Liang, Shuang and Shao, Jie and Ying, Rex},
  booktitle={Advances in Neural Information Processing Systems (NeurIPS), Datasets and Benchmarks Track},
  volume={37},
  year={2024}
}

@inproceedings{tolstikhin2021mlpmixer,
  title={{MLP-Mixer}: An all-{MLP} Architecture for Vision},
  author={Tolstikhin, Ilya O. and Houlsby, Neil and Kolesnikov, Alexander and Beyer, Lucas and Zhai, Xiaohua and Unterthiner, Thomas and Yung, Jessica and Steiner, Andreas and Keysers, Daniel and Uszkoreit, Jakob and Lucic, Mario and Dosovitskiy, Alexey},
  booktitle={Advances in Neural Information Processing Systems (NeurIPS)},
  volume={34},
  year={2021}
}
\end{document}